\pdfoutput=1
\documentclass[10pt,twocolumn,letterpaper]{article}

\usepackage[pagenumbers]{cvpr} 

\usepackage{amsmath,amssymb}
\usepackage{booktabs}
\usepackage{pifont}
\usepackage{colortbl}
\usepackage{xcolor}
\usepackage{multirow}
\usepackage{grffile}
\usepackage{stfloats}
\usepackage{caption}
\usepackage{microtype}
\usepackage{needspace}
\definecolor{cvprblue}{rgb}{0.21,0.49,0.74}
\usepackage[breaklinks,colorlinks,allcolors=cvprblue]{hyperref}

\def\confName{CVPR}
\def\confYear{2026}

\newcommand{\method}{GeoBoN}
\newcommand{\methodgated}{Gated GeoBoN}

\title{Test-Time Scaling for World Action Models via Zero-Shot Geometric Verification}

\author{
Zesen Zhao\textsuperscript{1} \quad
Minkyoung Cho\textsuperscript{1} \quad
Hui Shen\textsuperscript{1}\\
Boyuan Zheng\textsuperscript{1} \quad
Kunxiao Gao\textsuperscript{1} \quad
Yulong Cao\textsuperscript{2} \quad
Z. Morley Mao\textsuperscript{1}\\
\textsuperscript{1}University of Michigan \quad \textsuperscript{2}NVIDIA\\
{\tt\small \{hymanzzs, minkycho, huishen, boyuann, kunxiao, zmao\}@umich.edu}\\
{\tt\small yulongc@nvidia.com}
}

\begin{document}
\maketitle

\begin{abstract}
Test-time scaling improves foundation-model inference by spending additional computation, but robot control requires deciding whether extra compute is useful before executing an action. World Action Models (WAMs) make this decision natural: each rollout exposes both an action chunk and predicted future observations. We propose \methodgated, a training-free selective test-time scaling framework for WAMs. We first instantiate \method, a fixed-budget Best-of-$N$ selector that ranks sampled rollouts by cross-view depth reprojection consistency of their predicted futures, computed with a frozen geometry foundation model. \methodgated\ adds a lightweight action--future consistency gate that invokes \method\ only when the initial rollout appears internally inconsistent. Across five benchmark--backbone settings on RoboCasa, LIBERO Long, and RoboTwin~2.0, fixed-budget \method\ improves $N{=}8$ task success in every setting, e.g., raising the RoboCasa group average from $66.3\%$ to $68.4\%$ with Cosmos Policy and from $80.8\%$ to $82.5\%$ with X-WAM. With gating enabled, \methodgated\ recovers on average $74.8\%$ of the always-on success gain while triggering additional sampling on only $26.2\%$ of decision points. Offline diagnostics show that cross-view reprojection is a strong task-label-free selector, and we identify false low-score selections as a failure mode that helps explain why performance can saturate or degrade as $N$ increases.
\end{abstract}

\section{Introduction}
\label{sec:intro}

World Action Models (WAMs) have recently emerged as a promising paradigm for visuomotor control by jointly predicting future visual observations and action chunks from multi-view observations and language instructions~\citep{cosmos_policy,dreamzero,lingbotva}. Unlike action-only policies, a WAM exposes both a candidate action sequence and its predicted visual consequences before execution, creating an opportunity for test-time selection: the robot can inspect whether the imagined future is internally consistent, rather than comparing actions alone.

Test-time scaling has become an effective inference-time strategy in language models~\citep{cot,self_consistency,tree_of_thoughts,test_time_compute}, and robotic policies have begun to adopt a similar candidate-generation-and-selection paradigm using learned reward models, model-internal confidence signals, verifier objectives, or consistency criteria~\citep{robomonkey,rover,mgselect,wav,future_compatible}. However, these methods often either spend a fixed sampling budget at every decision point, or rely on selection signals that are trained, model-internal, or tied to a particular policy interface. This raises a basic question for WAM inference: when should the robot spend extra computation, and how should it choose among the sampled futures?

We propose a training-free two-stage consistency framework for selective test-time scaling of WAMs. The key idea is to audit the rollout using signals already exposed by WAM inference. A WAM rollout can fail along two axes: the generated action may disagree with the future imagined by the model, or the predicted futures from different camera views may fail to correspond to a single coherent 3D scene. These two checks have different costs. Action--future agreement can be tested cheaply from optical flow and projected end-effector motion, so we use it as a per-step gate. Cross-view geometric agreement requires running a geometry model on predicted frames, so we reserve it for selecting among candidates only when the cheap gate flags the initial rollout as unreliable. Concretely, if the gate finds the initial rollout consistent, the policy executes it directly; otherwise it samples additional rollouts, and a frozen VGGT-$\Omega$~\citep{wang2026vggtomega} geometry model ranks the candidates by a cross-view depth reprojection inconsistency score computed between the predicted primary-view and wrist-view future frames. Both stages operate directly on predicted images, generated actions, and proprioception, without task success labels, ground-truth future observations, online environment rollouts, or WAM-specific value heads.

In summary, we (i) identify cross-view geometric consistency as a training-free signal for ranking WAM rollouts and propose \textbf{\method}, a fixed-budget Best-of-$N$ selector; (ii) introduce an action--future consistency gate for selective test-time scaling, yielding \textbf{\methodgated}, which allocates additional sampling only when the initial rollout appears internally inconsistent; and (iii) validate the framework through closed-loop evaluations and offline diagnostics across multiple manipulation benchmarks and WAM backbones.

\begin{figure*}[t]
\centering
\includegraphics[width=\linewidth]{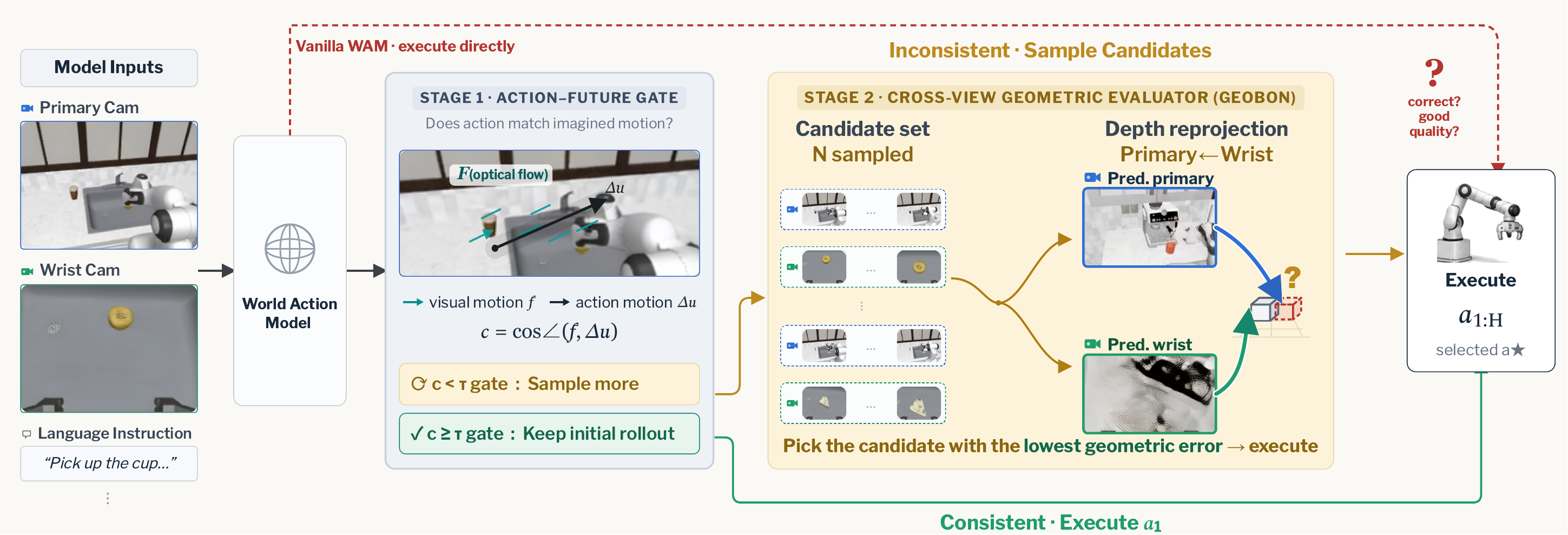}
\caption{The action--future gate decides whether the initial WAM rollout is reliable enough to execute directly; when it triggers, \method\ samples additional rollouts and selects the candidate with the lowest cross-view depth reprojection inconsistency.}
\label{fig:method}
\end{figure*}

\section{Related Work}
\label{sec:related}

\paragraph{World Action Models and imagined futures.}
WAMs jointly model robot actions and future observations, exposing an imagined rollout before execution. Cosmos Policy~\citep{cosmos_policy} and DreamZero~\citep{dreamzero} build on video-generation backbones, LingBotVA~\citep{lingbotva} studies causal video-action world modeling, and Motus~\citep{motus} and MotuBrain~\citep{motubrain} unify understanding, action, and video generation in latent-action world models. Recent 4D and video-action WAMs make future representations more explicit, including multiview RGB-D prediction in X-WAM~\citep{xwam} and unified action-conditioned evaluation in $\tau_0$-WM~\citep{tau0_wm}. These works improve the WAM backbone itself. Our work is orthogonal: given an existing multi-view WAM, we ask how its imagined futures can be used as training-free evidence for test-time selection.

\paragraph{Test-time scaling and rollout selection.}
Robotic policies increasingly adopt the sample-and-select paradigm of test-time scaling~\citep{cot,self_consistency,tree_of_thoughts,test_time_compute}: RoboMonkey uses a trained VLM-based verifier~\citep{robomonkey}, RoVer a robot process reward model~\citep{rover}, MG-Select model-internal masked-distribution signals~\citep{mgselect}, and WAV forward--inverse asymmetry of world-model predictions~\citep{wav}. For WAMs, concurrent work introduces future-consensus selection as a value-free fixed-budget selector~\citep{future_compatible}. These methods typically rely on a fixed sampling budget, a learned reward or verifier, or a model/interface-specific signal. In contrast, our method uses WAM-exposed consistency signals for selective inference: action--future consistency decides whether extra candidates are needed, and cross-view geometric consistency selects among them.

\paragraph{Adaptive and efficient WAM inference.}
A growing line of work reduces the cost of WAM-style future reasoning by questioning whether explicit imagination is always necessary~\citep{fastwam}, adapting the video denoising trajectory~\citep{sants}, comparing imagined and real observations after partial rollout~\citep{when_to_trust_imagination}, or via caching, compact backbones, and horizon-adaptive modeling~\citep{c3ache,efficient_wam,light_wam,aha_wam}. These methods mainly optimize the WAM architecture, denoising process, or execution horizon; our setting is complementary: before executing the current action chunk, we decide whether additional samples are worth drawing, and if so select among them with a frozen geometric evaluator.

\section{Method}
\label{sec:method}

WAM inference receives multi-view observations, proprioception, and a language instruction, and stochastically generates a visual--action rollout. We denote the $i$-th sampled rollout as
\begin{equation}
    \tau_i =
    \left(
    \hat{I}_{i}^{\mathrm{pri}},
    \hat{I}_{i}^{\mathrm{wri}},
    \mathbf{a}_{i,1:H}
    \right),
\end{equation}
where $\hat{I}_{i}^{\mathrm{pri}}$ and $\hat{I}_{i}^{\mathrm{wri}}$ are the predicted future frames used for scoring from the primary and wrist views, and $\mathbf{a}_{i,1:H}$ is the generated action chunk over horizon $H$. Our method uses two consistency signals: an action--future gate that decides whether additional candidates should be sampled, and a cross-view geometric evaluator that ranks candidates in the resulting candidate set (Figure~\ref{fig:method}).

\subsection{Action--Future Gate for Selective Sampling}
\label{sec:action_future_gate}

The gate tests whether the generated action is consistent with the motion visible in the predicted future. Given $\tau_1$, we compute dense optical flow $F_1$ between the current primary-view observation $I_t^{\mathrm{pri}}$ and the predicted future frame $\hat{I}_1^{\mathrm{pri}}$. For each arm $r$, let $\mathbf{x}_{r,0}$ be the current end-effector position from proprioception, and $\mathbf{x}_{1,r,H}$ the position after applying the generated action chunk (via forward kinematics for joint-position actions, or accumulated displacement for delta actions). Projecting both endpoints into the primary camera gives
\begin{equation}
    \Delta \mathbf{u}_{1,r}
    =
    \pi_{\mathrm{pri}}(\mathbf{x}_{1,r,H})
    -
    \pi_{\mathrm{pri}}(\mathbf{x}_{r,0}).
    \label{eq:projected_ee_motion}
\end{equation}

We ignore arms whose 3D displacement is below an idle threshold. For each moving arm, we average the optical flow inside a capsule-shaped region $\mathcal{M}_{1,r}$ around the projected end-effector trajectory:
\begin{equation}
    \bar{\mathbf{f}}_{1,r}
    =
    \frac{1}{|\mathcal{M}_{1,r}|}
    \sum_{p \in \mathcal{M}_{1,r}}
    F_1(p),
    \label{eq:arm_local_flow}
\end{equation}
and compute the cosine agreement between predicted visual motion and projected action motion:
\begin{equation}
    c_{1,r}
    =
    \frac{
    \bar{\mathbf{f}}_{1,r}^{\top} \Delta \mathbf{u}_{1,r}
    }{
    \|\bar{\mathbf{f}}_{1,r}\|_2
    \|\Delta \mathbf{u}_{1,r}\|_2
    + \epsilon
    }.
    \label{eq:flow_action_cosine}
\end{equation}

The gate triggers if any moving arm has $c_{1,r}<\tau_{\mathrm{gate}}$; otherwise, or if no arm remains after idle filtering, the policy executes the initial action chunk. When triggered, the policy samples $N_{\max}-1$ additional rollouts, retains the original in the candidate set, and ranks the candidates with the cross-view geometric evaluator.

\subsection{Cross-View Geometric Evaluator}
\label{sec:geo_evaluator}

For each candidate $\tau_i$, we feed the predicted frame pair $(\hat{I}_{i}^{\mathrm{pri}}, \hat{I}_{i}^{\mathrm{wri}})$ into frozen VGGT-$\Omega$. Let $d_i^{\mathrm{vggt}}(p)$ denote the depth predicted directly for the primary image at pixel $p$. Using the camera geometry estimated by VGGT-$\Omega$, we project the wrist-view 3D point map into the primary camera frame, yielding a projected depth $d_i^{\mathrm{proj}}(p)$. The cross-view depth reprojection inconsistency is
\begin{equation}
    e_{\mathrm{depth}}(\tau_i)
    =
    \frac{1}{|\Omega_i|}
    \sum_{p \in \Omega_i}
    \left|
    \log
    \frac{
    d_i^{\mathrm{proj}}(p)
    }{
    d_i^{\mathrm{vggt}}(p)
    }
    \right|,
    \label{eq:depth_consistency}
\end{equation}
where $\Omega_i$ contains pixels whose projected points fall inside the primary image with positive depths and VGGT-$\Omega$ confidence above a threshold $\gamma_{\mathrm{conf}}$; candidates with an empty valid set receive a large score. The logarithmic ratio reduces sensitivity to absolute depth scale. When selection is invoked, we execute the action chunk of the candidate with the lowest inconsistency.

\subsection{Inference Procedure}
\label{sec:inference_modes}

At each control step, the WAM first generates an initial rollout $\tau_1$. We instantiate three inference modes. The \emph{baseline} executes $\mathbf{a}_{1,1:H}$ directly, without additional sampling or geometric scoring. Fixed-budget \emph{\method} always samples a candidate set $\mathcal{C}_N=\{\tau_i\}_{i=1}^{N}$ and selects the rollout with the lowest cross-view reprojection inconsistency, $i^\star = \arg\min_{\tau_i \in \mathcal{C}_N} e_{\mathrm{depth}}(\tau_i)$, executing $\mathbf{a}_{i^\star,1:H}$. \emph{\methodgated} first applies the action--future gate from Sec.~\ref{sec:action_future_gate} to the initial rollout: if the gate does not trigger, it executes $\mathbf{a}_{1,1:H}$ directly; if it triggers, it samples $N_{\max}-1$ additional rollouts, forms $\mathcal{C}_{N_{\max}}$ including the initial rollout, and applies the same geometric selection rule.

\begin{table*}[t]
\centering
\scriptsize
\setlength{\tabcolsep}{3.0pt}
\renewcommand{\arraystretch}{1.08}
\caption{
\method\ success rate (\%). Bold indicates the best fixed-budget result in each row.
The last column reports the paired improvement of the main fixed-budget setting
($N=8$) over the baseline; brackets denote 95\% confidence intervals.
}
\label{tab:bon}
\vspace{1mm}
\resizebox{\linewidth}{!}{
\begin{tabular}{llcccccc}
\toprule
& & & \multicolumn{4}{c}{\textbf{\method}} & \textbf{$\Delta$ $N{=}8$ vs. Base (95\% CI)} \\
\cmidrule(lr){4-7}
\textbf{Benchmark / Task Group}
& \textbf{WAM}
& \textbf{Baseline}
& $N{=}2$
& $N{=}4$
& $N{=}8$
& $N{=}16$ \\
\midrule

\multicolumn{8}{l}{\textbf{RoboCasa}~\citep{robocasa}} \\
\quad Door/Drawer Manipulation & Cosmos Policy & 89.8 & 88.7 & 90.2 & 90.8 & \textbf{91.2} & $+1.0$ [$-2.6$, $+4.6$]\\
\quad Pick-and-Place           & Cosmos Policy & 47.2 & \textbf{49.2} & 47.9 & 49.0 & 48.1 & $+1.8$ [$-2.1$, $+7.7$]\\
\quad Appliance Control        & Cosmos Policy & 79.9 & 79.3 & 81.0 & \textbf{82.6} & 82.3 & $+2.7$ [$-2.0$, $+7.4$]\\
\quad Coffee Making            & Cosmos Policy & 38.3 & 38.7 & 38.5 & 42.0 & \textbf{42.9} & $+3.7$ [$-3.0$, $+13.0$] \\
\rowcolor{gray!12}
\quad \textbf{Group avg.}      & Cosmos Policy & 66.3 & 66.5 & 66.9 & \textbf{68.4} & 68.2 & $+2.1$ [$+0.6$, $+7.3$] \\

\midrule
\quad Door/Drawer Manipulation & X-WAM & 96.7 & 89.7 & \textbf{92.7} & 92.1 & 88.7 & $-4.6$ [$-6.9$, $-2.9$] \\
\quad Pick-and-Place           & X-WAM & 68.8 & 70.5 & 71.0 & 70.4 & \textbf{73.2} & $+1.6$ [$-6.6$, $+9.8$] \\
\quad Appliance Control        & X-WAM & 84.3 & 84.4 & 81.0 & \textbf{87.5} & 83.4 & $+3.2$ [$-2.3$, $+8.8$] \\
\quad Coffee Making            & X-WAM & 73.3 & 86.3 & 87.7 & 83.7 & \textbf{92.0} & $+10.4$ [$-2.8$, $+24.5$] \\
\rowcolor{gray!12}
\quad \textbf{Group avg.}      & X-WAM & 80.8 & 81.3 & 81.4 & \textbf{82.5} & 82.4 & $+1.7$ [$+0.3$, $+2.6$] \\

\midrule
\multicolumn{8}{l}{\textbf{LIBERO Long}~\citep{libero}} \\
& Cosmos Policy               & 97.5 & 98.2 & 98.6 & \textbf{99.3} & 98.9 & $+1.8$ [$+0.4$, $+2.2$] \\
& LingBotVA~\citep{lingbotva} & 97.2 & 97.8 & 98.0 & 98.3 & \textbf{99.1} & $+1.1$ [$+0.2$, $+1.5$] \\

\midrule
\multicolumn{8}{l}{\textbf{RoboTwin~2.0}~\citep{robotwin}} \\
& Motus~\citep{motus}         & 87.8 & 88.3 & 88.5 & \textbf{89.9} & 89.5 & $+2.1$ [$+0.2$, $+3.4$] \\

\bottomrule
\end{tabular}
}
\vspace{-1mm}
\end{table*}

\section{Experiments}
\label{sec:experiments}

We evaluate the framework along four axes: whether cross-view geometric evaluation improves task success under fixed-budget Best-of-$N$ inference; whether the action--future gate reduces sampling cost while preserving much of the always-on gain; whether reprojection and the gate are informative signals compared to alternatives; and why large Best-of-$N$ budgets can saturate or degrade under evaluator noise.

\subsection{Experimental Setup}
\label{sec:exp_setup}

We evaluate on RoboCasa~\citep{robocasa}, LIBERO Long~\citep{libero}, and RoboTwin~2.0~\citep{robotwin}, using Cosmos Policy~\citep{cosmos_policy}, X-WAM~\citep{xwam}, LingBotVA~\citep{lingbotva}, and Motus~\citep{motus} as WAM backbones. All experiments are training-free: we use publicly available checkpoints and their default evaluation setups without fine-tuning, on H200 and RTX Pro 6000 GPU nodes. We evaluate each benchmark--backbone setting with four random seeds, using the standard 50 rollouts per task per seed on LIBERO Long and 10 rollouts per task per seed on RoboCasa and RoboTwin~2.0; success rates are averaged over seeds. The gate uses Farneback optical flow~\citep{farneback2003twoframe} with idle threshold $\delta_{\mathrm{idle}}=1\mathrm{cm}$, cosine threshold $\tau_{\mathrm{gate}}=-0.2$, capsule radius $\rho_{\mathrm{cap}}=30$ pixels, and VGGT-$\Omega$ confidence threshold $\gamma_{\mathrm{conf}}=0.5$; all thresholds are fixed across experiments.

\subsection{Fixed-Budget \method}
\label{sec:exp_bon}

We compare single-rollout inference against fixed-budget \method\ with $N\in\{2,4,8,16\}$. Since X-WAM provides a native depth prediction head, X-WAM experiments use its own predicted depth for reprojection scoring rather than an external depth foundation model.

Across all benchmark--WAM settings in Table~\ref{tab:bon}, \method\ improves the main $N{=}8$ success rate over single-rollout inference. On RoboCasa, the task-count-weighted group average improves from $66.3\%$ to $68.4\%$ with Cosmos Policy and from $80.8\%$ to $82.5\%$ with X-WAM. Category-level results are mixed but informative: for X-WAM, Door/Drawer Manipulation drops from $96.7\%$ to $92.1\%$, while Coffee Making increases from $73.3\%$ to $83.7\%$ at $N{=}8$ and $92.0\%$ at $N{=}16$, suggesting that geometric reranking helps most when the task leaves room for rejecting implausible futures, but can hurt when the initial rollout is already strong. On LIBERO Long, \method\ improves Cosmos Policy from $97.5\%$ to $99.3\%$ and LingBotVA from $97.2\%$ to $98.3\%$ at $N{=}8$; on RoboTwin~2.0 with Motus, success improves from $87.8\%$ to $89.9\%$. Notably, several settings saturate or slightly degrade at $N{=}16$: fixed-budget geometric selection is not strictly monotonic in the number of sampled candidates.

\subsection{\methodgated}
\label{sec:exp_gated}

Table~\ref{tab:adaptive} evaluates the full gate-and-evaluator procedure. The goal of \methodgated\ is to recover most of the always-on \method\ gain while avoiding unnecessary candidate sampling and geometric scoring. Across the five benchmark--WAM settings, the gate triggers additional Best-of-$N$ sampling on only $14.2$--$34.5\%$ of decision points, while recovering $63.6$--$85.7\%$ of the always-on \method\ success gain.

\begin{table*}[t]
\centering
\scriptsize
\setlength{\tabcolsep}{4.0pt}
\renewcommand{\arraystretch}{1.08}
\caption{
\methodgated\ with $N_{\max}=8$. Full-gain recovery measures how much
of the always-on \method\ success gain is retained.
}
\label{tab:adaptive}
\vspace{1mm}
\resizebox{\linewidth}{!}{
\begin{tabular}{llcccc}
\toprule
\textbf{Benchmark}
& \textbf{Mode}
& \textbf{Success Rate}
& \textbf{Full-Gain Recovery}
& \textbf{BoN Trigger (\%)}
& \textbf{Avg. Latency (s)} \\
\midrule

\multicolumn{6}{l}{\textbf{RoboCasa} \quad / \quad Cosmos Policy} \\
& Baseline
& 66.3\% & 0.0\% & 0\% & 0.90 \\
\rowcolor{gray!10}
& \methodgated\ ($N{=}8$)
& 67.9\% & 76.2\% & 24.7\% & 1.29 \\
& \method\ ($N{=}8$)
& 68.4\% & 100.0\% & 100\% & 3.65 \\

\midrule

\multicolumn{6}{l}{\textbf{RoboCasa}\quad / \quad X-WAM} \\
& Baseline
& 80.8\% & 0.0\% & 0\% & 2.68 \\
\rowcolor{gray!10}
& \methodgated\ ($N{=}8$)
& 82.1\% & 76.5\% & 25.2\% & 3.11 \\
& \method\ ($N{=}8$)
& 82.5\% & 100.0\% & 100\% & 9.67 \\

\midrule

\multicolumn{6}{l}{\textbf{LIBERO Long} \quad / \quad Cosmos Policy} \\
& Baseline
& 97.5\% & 0.0\% & 0\% & 0.90 \\
\rowcolor{gray!10}
& \methodgated\ ($N{=}8$)
& 98.8\% & 72.2\% & 14.2\% & 0.96 \\
& \method\ ($N{=}8$)
& 99.3\% & 100.0\% & 100\% & 3.66 \\

\midrule

\multicolumn{6}{l}{\textbf{LIBERO Long} \quad / \quad LingBotVA} \\
& Baseline
& 97.2\% & 0.0\% & 0\% & 2.27 \\
\rowcolor{gray!10}
& \methodgated\ ($N{=}8$)
& 97.9\% & 63.6\% & 34.5\% & 2.83 \\
& \method\ ($N{=}8$)
& 98.3\% & 100.0\% & 100\% & 3.86 \\

\midrule

\multicolumn{6}{l}{\textbf{RoboTwin~2.0} \quad / \quad Motus} \\
& Baseline
& 87.8\% & 0.0\% & 0\% & 2.29 \\
\rowcolor{gray!10}
& \methodgated\ ($N{=}8$)
& 89.6\% & 85.7\% & 32.2\% & 2.83 \\
& \method\ ($N{=}8$)
& 89.9\% & 100.0\% & 100\% & 3.86 \\

\bottomrule
\end{tabular}
}
\vspace{-1mm}
\end{table*}

\methodgated\ closely tracks the fixed-budget result at much lower latency. On RoboCasa, it improves success from $66.3\%$ to $67.9\%$ (Cosmos Policy) and from $80.8\%$ to $82.1\%$ (X-WAM) at roughly one-third of the always-on latency; on LIBERO Long with Cosmos Policy it reaches $98.8\%$ with only a $14.2\%$ trigger rate, and on RoboTwin~2.0 with Motus $89.6\%$, close to the always-on $89.9\%$. Overall, gating preserves most of the fixed-budget benefit at a fraction of the average cost.

\subsection{Selector and Gate Diagnostics}
\label{sec:diagnostics}

We ablate the two decisions made by our system: which rollout to select, and when to invoke additional sampling. Offline diagnostics use the same fixed $N=8$ candidate dumps.

\begin{table*}[t]
\centering
\small
\setlength{\tabcolsep}{4pt}
\renewcommand{\arraystretch}{1.05}
\caption{
Selector evaluation using offline error recovery $\mathrm{ER}$ (\%) and online closed-loop success rate (\%). ER is computed from fixed $N=8$ candidate dumps; the offline baseline is set to $0$ by definition. Higher is better; best per row in each block is shown in \textbf{bold}.
}
\label{tab:selection_combined}
\resizebox{\textwidth}{!}{
\begin{tabular}{lrrrrcrrrr}
\toprule
& \multicolumn{4}{c}{\textbf{Error Recovery (\%)}}
& & \multicolumn{4}{c}{\textbf{Online success (\%)}}\\
\cmidrule(lr){2-5} \cmidrule(lr){7-10}
\textbf{Benchmark--WAM}
& Baseline & VGGT Conf. & Consensus & \method
& & Baseline & VGGT Conf. & Consensus & \method \\
\midrule

RoboCasa / Cosmos Policy
& 0 & 7.3 & 6.9 & \textbf{7.6} & & 66.3 & 65.9 & 67.1 & \textbf{68.4} \\

RoboCasa / X-WAM
& 0 & 4.0 & \textbf{8.6} & 6.9 & & 80.8 & 77.5 & 77.4 & \textbf{82.5} \\

LIBERO Long / LingBotVA
& 0 & 11.4& 21.3& \textbf{22.0} & & 97.2 & 96.3 & 95.3 & \textbf{98.3} \\

RoboTwin~2.0 / Motus
& 0 & 1.5 & 7.6 & \textbf{8.3} & & 87.8 & 88.1 & \textbf{90.5}& 89.9 \\

\bottomrule
\end{tabular}
}
\vspace{-1mm}
\end{table*}

\paragraph{Selector ablation.}
Table~\ref{tab:selection_combined} compares our reprojection selector with two training-free alternatives: a VGGT-$\Omega$ confidence-based selector and a future-consensus selector adapted from~\cite{future_compatible} (reimplemented from the paper description, as the original implementation is not public). We evaluate each selector with offline error recovery (ER) and online closed-loop success. ER measures the fraction of the gap closed between the default first-candidate selector and a ground-truth-aware oracle, $\mathrm{ER} = (\mathrm{ADE}_{c_0}-\mathrm{ADE}_{s})/(\mathrm{ADE}_{c_0}-\mathrm{ADE}_{\mathrm{oracle}}) \times 100$; the oracle is used only for offline analysis.

Cross-view reprojection is the most consistent selector among the tested signals: it achieves the best offline ER in three of four settings, and is the only selector that improves online success over the baseline in all settings. Confidence-only ranking is unstable and often reduces online success, and consensus, while competitive in individual cases, is less consistent in closed loop. These results support reprojection inconsistency as the second-stage selector, while also showing that offline error recovery is only an approximate proxy for closed-loop success.

\begin{table}[t]
\centering
\scriptsize
\setlength{\tabcolsep}{3.2pt}
\renewcommand{\arraystretch}{1.05}
\caption{Equal-budget gate diagnostic on fixed $N=8$ candidate dumps.
Random uses the same trigger rate.}
\label{tab:gate_diagnostic}
\vspace{1mm}
\resizebox{\columnwidth}{!}{
\begin{tabular}{lcccc}
\toprule
\textbf{Benchmark--WAM}
& \textbf{Trigger}
& \multicolumn{3}{c}{\textbf{\method\ Help Rate (\%)}} \\
\cmidrule(lr){3-5}
& \textbf{(\%)} & \textbf{Random} & \textbf{AF Gate} & $\boldsymbol{\Delta}$ \\
\midrule
RoboCasa/X-WAM         & 25.2 & 43.2 & 67.2 & $+24.0$ \\
RoboCasa/Cosmos Policy & 24.7 & 43.2 & 63.5 & $+20.3$ \\
LIBERO/LingBotVA       & 34.5 & 47.8 & 69.4 & $+21.6$ \\
RoboTwin/Motus         & 32.2 & 71.7 & 75.8 & $+4.1$ \\
\bottomrule
\end{tabular}
}
\vspace{-1mm}
\end{table}

\paragraph{Gate ablation.}
We next test whether the gate provides an informative compute-allocation signal rather than merely reducing the sampling budget. Defining $g=\mathrm{ADE}_{c_0}-\mathrm{ADE}_{\mathrm{GeoBoN}}$ so that $g>0$ means invoking \method\ helps, Table~\ref{tab:gate_diagnostic} compares the gate with a random trigger at the same trigger rate. The gate improves the \method\ help rate in all settings, with large gains on RoboCasa/X-WAM ($67.2\%$ vs. $43.2\%$), RoboCasa/Cosmos Policy ($63.5\%$ vs. $43.2\%$), and LIBERO/LingBotVA ($69.4\%$ vs. $47.8\%$), and a smaller gain on RoboTwin/Motus ($75.8\%$ vs. $71.7\%$): the signal is informative but not uniformly strong across domains.

\begin{figure}[t]
\centering
\includegraphics[width=0.95\linewidth]{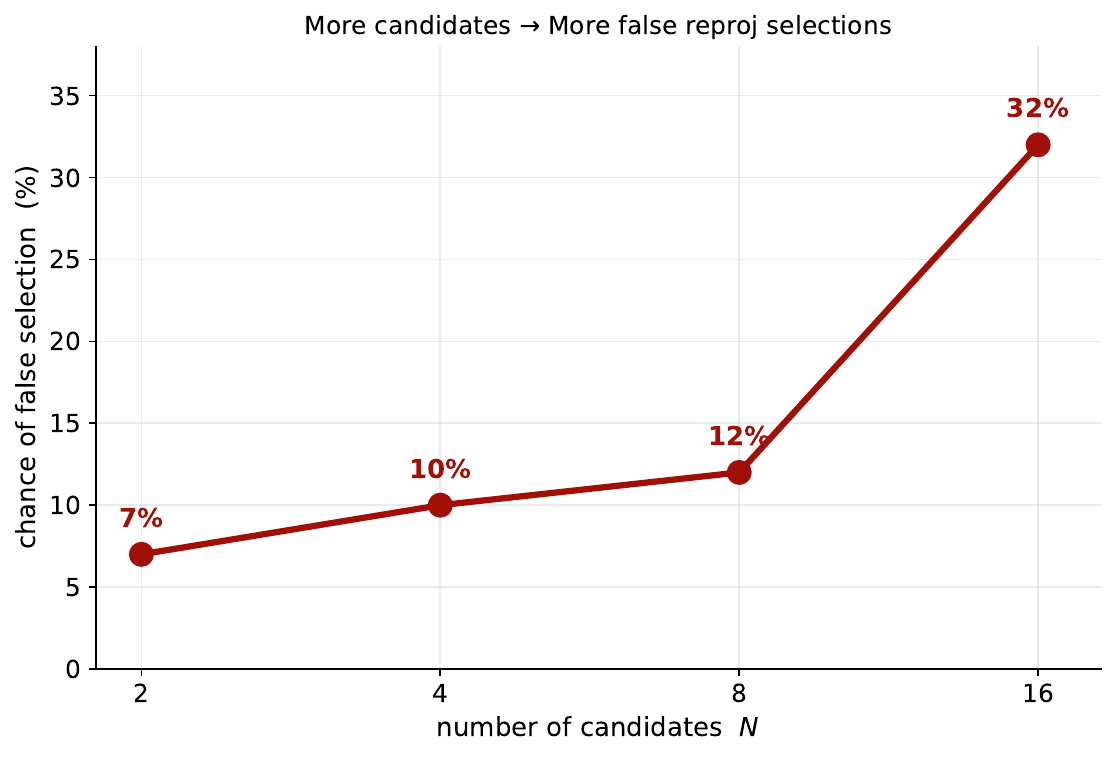}
\caption{False low-score selection rate as the candidate budget increases.
Larger candidate pools create more opportunities for Best-of-$N$ to select
a spurious low-reprojection outlier.}
\label{fig:hack_rate}
\vspace{-2mm}
\end{figure}

\begin{figure*}[b]
\centering
\includegraphics[width=0.85\linewidth]{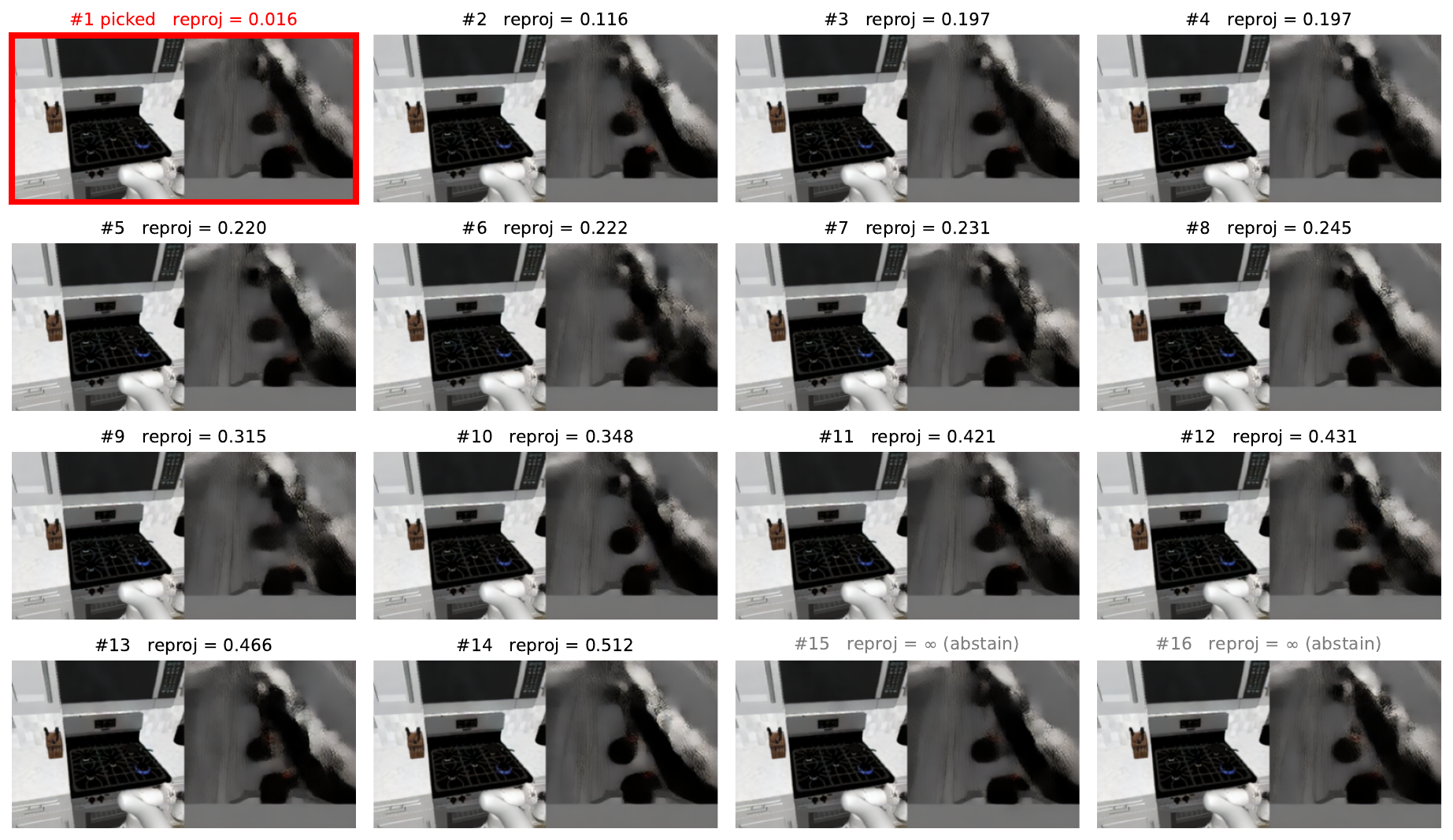}
\caption{Representative false low-score selection. The candidate futures are visually
near-identical, but the geometric evaluator assigns one candidate an unusually
low reprojection error and Best-of-$N$ selects it.}
\label{fig:hack}
\vspace{-2mm}
\end{figure*}

\paragraph{Comparison to a model-specific value head.}
We also compare \method\ with the Cosmos Policy value head, a model-specific learned selector, using identical candidate budgets. As shown in Table~\ref{tab:value_head}, \method\ consistently outperforms the value head across all budgets: by $+1.3$ to $+2.3$ percentage points on RoboCasa and $+0.7$ to $+1.3$ points on near-saturated LIBERO Long. Cross-view geometric consistency thus serves as a strong training-free alternative to model-internal value estimates, while remaining applicable to backbones without a learned value head.

\begin{table}[t]
\centering
\scriptsize
\setlength{\tabcolsep}{3.2pt}
\renewcommand{\arraystretch}{1.05}
\caption{
Comparison with the Cosmos Policy value head.
Both methods use the same candidate budget; only the selector changes.
}
\label{tab:value_head}
\vspace{1mm}
\resizebox{\columnwidth}{!}{
\begin{tabular}{lccccc}
\toprule
\textbf{Benchmark}
& \textbf{Selector}
& $N{=}2$
& $N{=}4$
& $N{=}8$
& $N{=}16$ \\
\midrule
RoboCasa
& Value Head & 65.2 & 65.5 & 66.1 & 65.9 \\
& \method     & 66.5 & 66.9 & 68.4 & 68.2 \\
\midrule
LIBERO Long
& Value Head & 97.1 & 97.9 & 98.1 & 97.6 \\
& \method     & 98.2 & 98.6 & 99.3 & 98.9 \\
\bottomrule
\end{tabular}
}
\vspace{-1mm}
\end{table}

\subsection{Best-of-$N$ Failure Analysis}
\label{sec:failure}

We further analyze why fixed-budget \method\ does not always improve with the candidate budget. A desirable evaluator should assign similar scores to visually near-identical futures, so we look for \emph{false low-score selections}: cases where Best-of-$N$ selects a candidate with an unusually low reprojection error despite a near-duplicate rendering with no visible geometric difference. We identify near-duplicates via LPIPS~\citep{lpips} distance below $0.02$, and mark a selection as false when its score is separated from a near-duplicate by more than the evaluator's 95th-percentile duplicate-pair variation.

Figure~\ref{fig:hack} shows a representative case: the candidate futures are visually near-identical, but the evaluator assigns one a much lower reprojection error and Best-of-$N$ selects it. This failure mode grows with the candidate pool: the false low-score selection rate increases from $7\%$ at $N{=}2$ to $12\%$ at $N{=}8$ and $32\%$ at $N{=}16$ (Fig.~\ref{fig:hack_rate}). This suggests a multiple-comparisons effect: larger pools raise the chance of finding a genuinely better rollout, but also of selecting a spurious outlier, explaining why \method\ saturates at larger $N$ and motivating moderate budgets with selective invocation.

\section{Conclusion}
\label{sec:conclusion}
\enlargethispage{\baselineskip}

\looseness=-1
We presented \methodgated, a training-free selective test-time scaling framework for World Action Models that uses the futures already exposed by WAM rollouts for inference-time decisions: action--future consistency decides when additional sampling is worth invoking, and cross-view geometric consistency ranks the sampled candidates. Across RoboCasa, LIBERO Long, and RoboTwin~2.0, fixed-budget \method\ improves Best-of-$N$ rollout selection across multiple WAM backbones, while \methodgated\ recovers much of the always-on gain at a substantially smaller sampling budget. Our diagnostics show that cross-view reprojection is a more consistent task-label-free selector than confidence-only scoring, that the gate provides an informative selective-compute signal, and that false low-score selections explain saturation at large $N$. These findings suggest that WAMs benefit not only from sampling more futures, but from using them to decide when more computation is needed.

\clearpage
{
    \small
    \bibliographystyle{ieeenat_fullname}
    \bibliography{main}
}

\end{document}